\documentclass[runningheads]{llncs}
\usepackage[T1]{fontenc}
\usepackage{graphicx}
\usepackage[utf8]{inputenc} % allow utf-8 input
\usepackage[T1]{fontenc}    % use 8-bit T1 fonts
\usepackage{hyperref}       % hyperlinks
\usepackage{url}            % simple URL typesetting
\usepackage{booktabs}       % professional-quality tables
\usepackage{amsfonts}       % blackboard math symbols
\usepackage{nicefrac}       % compact symbols for 1/2, etc.
\usepackage{microtype}      % microtypography
\usepackage{xcolor}         % colors
\usepackage{microtype}
\usepackage{graphicx}
\usepackage{subcaption,wrapfig}
\usepackage{booktabs} % for professional tables
\usepackage{tikz}
\usetikzlibrary{spy}
\usepackage{multirow, makecell,enumitem}
\usepackage{amssymb}
\usepackage{mathtools}
\usepackage[capitalize,noabbrev]{cleveref}
\usepackage{amsmath}

\newcommand{\norm}[1]{\left|\left|{#1}\right|\right|}
\newcommand{\prox}{\operatorname{Prox}}

\newcommand{\id}{\operatorname{Id}}

\renewcommand{\Im}{\operatorname{Im}}

\newcommand{\dime}{n}

\DeclareMathOperator*{\argmin}{arg\,min}

\newcommand{\reglambda }{\lambda}
\usepackage[textsize=tiny]{todonotes}
\usepackage{amsfonts}
\usepackage[overload]{empheq}

\begin{document}

\title{A relaxed proximal gradient descent algorithm for convergent Plug-and-Play %image restoration 
with proximal denoiser}

\titlerunning{Relaxed Proximal Gradient Descent for Convergent Plug-and-Play}
\author{Samuel Hurault\inst{1}, Antonin Chambolle\inst{2}, Arthur Leclaire\inst{1}, Nicolas Papadakis\inst{1}
}
\authorrunning{S. Hurault et al.%F. Author et al.
}
\institute{Univ. Bordeaux, CNRS, Bordeaux INP, IMB, UMR 5251, F-33400 Talence, France \and
CEREMADE, CNRS, Universit\'e Paris-Dauphine, PSL, Paris 75775, France}

\maketitle              % typeset the header of the contribution
\begin{abstract}
This paper presents a new convergent Plug-and-Play (PnP) algorithm.
PnP methods are efficient iterative algorithms for solving image inverse problems formulated as the minimization of the sum of a data-fidelity term and a regularization term. PnP methods perform regularization by plugging a pre-trained denoiser in a proximal algorithm, such as Proximal Gradient Descent (PGD). To ensure convergence of~PnP schemes, many  works study specific parametrizations of deep denoisers. However, existing results require either unverifiable or suboptimal hypotheses on the denoiser, or assume  restrictive conditions on the parameters of the inverse problem. Observing that these limitations can be due to the proximal algorithm in use, we study a relaxed version of the PGD algorithm for minimizing the sum of a convex function and a weakly convex one. When plugged with a relaxed proximal denoiser, we show that the proposed PnP-$\alpha$PGD algorithm converges for a wider range of regularization parameters, thus allowing more accurate image restoration.

\end{abstract}
\section{Introduction}

We focus on the convergence of plug-and-play methods associated to the class of inverse problems:
\begin{equation}
    \label{eq:pb}
    \hat x \in \argmin_{x} \lambda f(x) + \phi(x),
\end{equation}
where $f$ is a $L_f$-Lipschitz gradient function acting as a data-fidelity term with respect to a degraded  observation $y$,  $\phi$ is a $M$-weakly convex regularization function (such that $\phi+ \frac{M}{2}\norm{.}^2$ is convex); and $\lambda>0$ is a parameter weighting the influence of both terms. 
In our experimental setting, we consider degradation models $y=Ax^*+\nu\in \mathbb{R}^m$ for some groundtruth signal $x^*\in\mathbb{R}^n$,  a linear operator $A\in\mathbb{R}^{n\times m}$ and a white Gaussian noise $\nu\in \mathbb{R}^n$. We thus deal with convex data-fidelity terms of the form  $f(x)=\frac12||Ax-y||^2$, with $L_f=||A^T A||$. Our analysis can nevertheless apply to a broader class of convex or nonconvex functions $f$.

To find an adequate solution of the ill-posed problem of recovering $x^*$ from~$y$, the choice of the regularization $\phi$ is crucial. 
Convex~\cite{ROF} and nonconvex~\cite{wen2018survey} handcrafted  functions are now largely outperformed by learning approaches~\cite{romano2017little,zhang2021plug}, that may not even be associated to a closed-form regularization function $\phi$.

\subsection{Proximal algorithms} 
Estimating a local or global optima of problem~\eqref{eq:pb} is classically done using proximal splitting algorithms such as Proximal Gradient Descent~(PGD) or Douglas-Rashford Splitting (DRS).
Given \hbox{an adequate stepsize $\tau>0$,} these methods alternate between explicit gradient descent, $\id-\tau \nabla h$ for   smooth functions $h$, and/or implicit gradient steps using the proximal operator
$\prox_{\tau h}(x) \in \argmin_z \frac1{2\tau}||z-x||^2+h(z)$ , for proper lower semi-continuous functions $h$.
Proximal algorithms are originally designed for convex functions, but under appropriate assumptions, PGD~\cite{attouch2013convergence} and DRS~\cite{themelis2020douglas}  algorithms converge to a stationary point of problem~\eqref{eq:pb} associated to nonconvex functions $f$ and $\phi$.

\subsection{Plug-and-Play algorithms} 
Plug-and-Play (PnP)~\cite{venkatakrishnan2013plug} and Regularization by Denoising (RED)~\cite{romano2017little} methods consist in splitting algorithms in which the descent step on the regularization function is performed by an off-the-shelf image denoiser. They are respectively built from proximal splitting schemes by replacing the proximal operator (PnP) or the gradient operator (RED) of the regularization $\phi$ by an image denoiser. 
When used with a deep denoiser (\emph{i.e} parameterized by a neural network) these approaches produce impressive  results for various image restoration tasks~\cite{zhang2021plug}. 

Theoretical convergence of PnP and RED algorithms with deep denoisers has recently been addressed by a variety of studies~\cite{ryu2019plug,sun2021scalable,pesquet2021learning,hertrich2021convolutional}. Most of these works require specific constraints on the deep denoiser, such as nonexpansivity. 
However, imposing nonexpansivity of a denoiser can severely degrade its denoising performance.

Another line of works tries to address convergence by making PnP and RED algorithms be exact proximal algorithms. The idea is to replace the denoiser of RED algorithms by a gradient descent operator and the one of PnP algorithm by a proximal operator. Theoretical convergence then follows from known convergence results of proximal splitting algorithms. The authors of~\cite{cohen2021has,hurault2021gradient} thus propose to plug an explicit \emph{gradient-step denoiser} of the form $D=\id-\nabla g$, for a tractable and potentially nonconvex potential $g$ parameterized by a neural network. 
As shown in~\cite{hurault2021gradient}, such a constrained parametrization does not harm denoising performance. 
The gradient-step denoiser guarantees convergence of RED methods without sacrificing performance, but it does not cover convergence of PnP algorithms. 
An extension to PnP has been addressed in~\cite{hurault2022proximal}: following~\cite{gribonval2020characterization}, when $g$ is trained with contractive gradient, the gradient-step denoiser can be written as a proximal operator $D=\id-\nabla g=\prox_{\phi}$ of a nonconvex potential $\phi$.
A PnP scheme with this \emph{proximal denoiser} becomes again a genuine proximal splitting algorithm associated to an explicit functional. Following existing convergence results of the PGD and DRS algorithms in the nonconvex setting, \cite{hurault2022proximal} proves convergence of PnP-PGD and PnP-DRS with proximal denoiser. 

The main limitation of this approach is that the proximal denoiser $D=\prox_{\phi}$ does not give tractability of $\prox_{\tau \phi}$ for $\tau \neq 1$.
Therefore, to be a provable converging proximal splitting algorithm, the stepsize of the overall PnP algorithm has to be fixed to $\tau=1$. 
For instance, for the PGD algorithm with stepsize $\tau=1$:
\begin{equation}
    \label{eq:PGD0}
    x_{k+1} \in \prox_{\phi} (\id - \lambda \nabla f)(x_k)
\end{equation}
the convergence of ${x_k}$ to a stationary point of~\eqref{eq:pb} is only ensured for regularization parameters $\lambda$ satisfying $L_f\lambda<1$~\cite{attouch2013convergence}. This is an issue for low noise levels $\nu$, for which relevant solutions are obtained with a dominant data-fidelity term in~\eqref{eq:pb}. 

Our objective is to design a convergent PnP algorithm with a proximal denoiser, and with minimal restriction on the regularization parameter $\lambda$. Contrary to previous work on PnP convergence~\cite{ryu2019plug,sun2021scalable,terris2020building,cohen2021has,hurault2021gradient,hertrich2021convolutional}, we not only wish to adapt the denoiser but also the original optimization scheme of interest. We study a new proximal algorithm able to deal with a proximal operator that can only be computed for a predefined and fixed stepsize $\tau=1$.

\subsection{Contributions and outline} In this paper, we propose a  relaxation of the Proximal Gradient Descent algorithm, called \emph{$\alpha$PGD}, such that when used with a proximal denoiser, the corresponding PnP scheme \emph{Prox-PnP-$\alpha$PGD} can converge for any regularization parameter $\lambda$.  

In section~\ref{sec2}, extending the result from \cite{hurault2022proximal}, we show how building a denoiser $D$ that corresponds to the proximal operator of a $M$-weakly convex function $\phi$. 
Then we introduce a relaxation of the denoiser that allows to control $M$, the constant of weak convexity of $\phi$.

In section~\ref{sec3}, we give new results on the convergence of Prox-PnP-PGD \cite{hurault2022proximal} with regularization constraint $\lambda(L_f+M)<2$.  In particular, using the convergence of PGD for nonconvex $f$ and weakly convex $\phi$ given in Theorem~\ref{prop:PGD}, Corollary~\ref{thm:PnP-PGD} improves previous PnP convergence results~\cite{hurault2022proximal} for $M<L_f$.

In section~\ref{sec4}, we present \emph{$\alpha$PGD}, a relaxed version of the PGD algorithm\footnote{There are two different notions of relaxation in this paper. One is for the relaxation of the proximal denoiser and the other for the relaxation of the optimization scheme.} reminiscent of the accelerated PGD scheme from~\cite{tseng2008accelerated}. Its convergence is shown Theorem~\ref{thm:alphaPGD} 
for a smooth convex function~$f$ and a weakly convex one $\phi$.
Corollary~\ref{thm:PnP-PGD2} then illustrates how the relaxation parameter $\alpha$ can be tuned to make the proposed PnP-$\alpha$PGD algorithm convergent for regularization parameters $\lambda$ satisfying $\lambda L_f M <1$. Having a multiplication, instead of an addition, between the constants $L_f$ and $M$ opens new perspectives. 
In particular, by plugging a relaxed denoiser with controllable weak convexity constant, 
Corollary~\ref{thm:PnP-PGD3} demonstrates that, for all  regularization parameter $\lambda$, we can always decrease $M$ such that $\lambda L_f M <1$ i.e. such that the $\alpha$PGD algorithm converges.
 
In section 5, we provide experiments for both image deblurring and image super-resolution applications. We demonstrate the effectiveness of our PnP-$\alpha$PGD algorithm, which closes the performance gap between Prox-PnP-PGD and the state-of-the-art plug-and-play algorithms.

\section{Relaxed Proximal Denoiser}\label{sec2}

This section introduces the denoiser used in our PnP algorithm. We first redefine the Gradient Step denoiser and show in Proposition~\ref{prop:proxdenoiser} how it can be constrained to be a proximal denoiser; and finally introduced the relaxed proximal denoiser.

\subsection{Gradient Step  Denoiser}

In this paper, we use the Gradient Step Denoiser introduced in~\cite{hurault2021gradient,cohen2021has}. 

It is defined as a gradient step over a differentiable potential~$g_\sigma$ parametrized by a neural network. 
\begin{equation}
    \label{eq:gs_denoiser}
    D_\sigma = \id - \nabla g_\sigma.
\end{equation}

This denoiser can then be trained to denoise white Gaussian noise $\nu_\sigma$ of various standard deviations $\sigma$ by minimizing the $\ell_2$ denoising loss $\mathbf{E}[||D_\sigma(x+\nu_\sigma)-x)||^2]$. When parametrized using a DRUNet  architecture~\cite{zhang2021plug}, it was shown in~\cite{hurault2021gradient} that the Gradient Step Denoiser~\eqref{eq:gs_denoiser}, despite being constrained to be a conservative vector field (as in~\cite{romano2017little}), achieves state-of-the-art denoising performance.

\subsection{Proximal Denoiser}
We propose here a new version of the result of~\cite{hurault2022proximal} on the characterization of the gradient-step denoiser as a proximal operator of some potential $\phi$. In particular, we state a new result regarding the weak convexity of the $\phi$ function. The proof of this result, given in Appendix~\ref{app:proxdenoiser} relies on the results from~\cite{gribonval2020characterization}. 

\begin{proposition}[Proximal denoisers]
    \label{prop:proxdenoiser} Let ${g_\sigma : \mathbb{R}^n \to \mathbb{R}}$ a $\mathcal{C}^{2}$ function with $\nabla g_\sigma$ $L_{g_\sigma}$-Lipschitz with ${L_{g_\sigma}<1}$. Then, for ${D_\sigma :=  \id - \nabla g_\sigma}$, there exists a potential ${\phi_\sigma:\mathbb{R}^n \to \mathbb{R} \cup \{+\infty\}}$ such that $\prox_{\phi_\sigma}$ is one-to-one and 
    \begin{equation}
    \label{eq:proximal_denoiser}
        D_\sigma = \prox_{\phi_\sigma}
    \end{equation}
    Moreover, 
   $\phi_\sigma$ is $\frac{L_{g_\sigma}}{L_{g_\sigma}+1}$-weakly convex %
   and it can be written 
$\phi_\sigma = \hat \phi_\sigma + K$ on $\Im(D_\sigma)$ (which is open) for some constant $K\in \mathbb{R}$,  with ${\hat \phi_\sigma : \mathcal{X} \to \mathbb{R} \cup \{+\infty\}}$ defined by
        \begin{equation}
            \label{eq:phi}
            \hat \phi_\sigma(x):=\left\{\begin{array}{ll}   g_\sigma({D_\sigma}^{-1}(x)))-\frac{1}{2} \norm{{D_\sigma}^{-1}(x)-x}^2  \ \  \text{if }\ x \in \Im(D_\sigma),\\
             +\infty \ \ \text{otherwise.} \end{array}\right.
        \end{equation} 
    Additionally $\hat \phi_\sigma$ verifies $\forall x \in \mathbb{R}^n$,  $\hat \phi_\sigma(x) \geq g_\sigma(x)$. 
\end{proposition}
To get a proximal denoiser from the  denoiser~\eqref{eq:gs_denoiser}, the gradient of the learned potential $g_\sigma$ must be contractive.
In~\cite{hurault2022proximal} the Lipschitz constant of  $\nabla g_\sigma$ is softly constrained to satisfy $L_{g_\sigma}<1$, by penalizing the spectral norm $\norm{\nabla^2 g_\sigma(x+\nu_\sigma)}_S$ in the denoiser training loss. 

\subsection{Relaxed Denoiser}

Once trained,  the Gradient Step Denoiser ${D_\sigma = \id - \nabla g_\sigma}$ can be relaxed as in~\cite{hurault2021gradient} with a parameter $\gamma \in [0,1]$
\begin{equation}
    \label{eq:relaxed_proximal_denoiser}
    D^\gamma_\sigma = \gamma D_\sigma + (1-\gamma)\id = \id - \gamma \nabla g_\sigma.
\end{equation}
Applying Proposition~\ref{prop:proxdenoiser} with $g^\gamma_\sigma = \gamma g_\sigma$ which has a $\gamma L_{g_\sigma}$-Lipschitz gradient, we get that if $\gamma L_g < 1$, there exists a $\frac{\gamma L_{g_\sigma}}{\gamma L_{g_\sigma}+1}$-weakly convex ${\phi^\gamma_\sigma
}$ 
such that
\begin{equation}\label{eq:relaxed_proximal_denoiser_phi}
D^\gamma_\sigma = \prox_{\phi^\gamma_\sigma},
\end{equation}
satisfying $\phi^0_\sigma=0$ and $\phi^1_\sigma=\phi_\sigma$. 
Hence, one can control the weak convexity of the regularization function by relaxing the proximal denoising operator $D^\gamma_\sigma$.

\section{Plug-and-Play Proximal Gradient Descent (PnP-PGD)}\label{sec3}

In this section, we give convergence results for the  Prox-PnP-PGD algorithm, $x_{k+1} = D_\sigma \circ (\id - \lambda f)(x_k) = \prox_{\phi_{\sigma}} \circ (\id - \lambda f)(x_k)$, 
which is the PnP version of PGD, with plugged Proximal Denoiser \eqref{eq:proximal_denoiser}. The authors of \cite{hurault2022proximal} proposed a suboptimal convergence result as the semiconvexity of $\phi_\sigma$ was not exploited. We here improve the condition on the regularization parameter $\lambda$ for convergence.

We present properties of smooth functions and weakly convex functions in Section~\ref{sec:pgd_prop}.
Then we show in Section~\ref{sec:pgd} the convergence of PGD in the smooth/weakly convex setting. We finally apply this result to PnP in Section~\ref{sec:pgd_prox}. 

\subsection{Useful inequalities}\label{sec:pgd_prop}
We present two results relative to weakly convex functions and smooth ones. We use the subdifferential of a proper, nonconvex function $\phi$  defined as 
$\partial \phi(x)\hspace{-2pt}=\hspace{-2pt}\left\{ 
v \hspace{-1pt}\in \hspace{-1pt}\mathbb{R}^n\hspace{-1pt}, 
\exists (x_k), \phi(x_k) \hspace{-1.5pt}\rightarrow\hspace{-1.5pt} \phi(x), v_k\hspace{-1.5pt} \rightarrow\hspace{-1.5pt} v,
\varliminf_{z \to x_k} \hspace{-7pt}\frac{\phi(z)-\phi(x_k) - \langle v_k,z-x_k \rangle }{\norm{z-x_k}}\hspace{-2pt} \geq \hspace{-2pt} 0 \ \forall k \hspace{-1pt}\right\}\hspace{-2pt}.$

\begin{proposition}[Properties of weakly convex functions, proof in Appendix~\ref{app:3points}]
\label{prop:weaklyconvex}
For $\phi$ proper lsc and $M$-weakly convex with $M>0$,
\begin{itemize}
    \item[(i)]  $\forall x,y$ and $t \in [0,1]$, 
    \begin{equation} 
    \phi(tx+(1-t)y) \leq t\phi(x) + (1-t)\phi(y) + \frac{M}{2}t(1-t)\norm{x-y}^2;
    \end{equation}
    \item[(ii)] $\forall x,y$, we have $\forall z \in \partial \phi(y)$, 
    \begin{equation}
    \phi(x) \geq \phi(y) + \langle z,x-y \rangle - \frac{M}{2}\norm{x-y}^2;
\end{equation}
 \item[(iii)] \textbf{Three-points inequality}. 
For $z^+ \in \prox_\phi(z) $, we have, $\forall x$
\begin{equation}
    \phi(x) + \frac{1}{2}\norm{x-z}^2 \geq \phi(z^+) +  \frac{1}{2}\norm{z^+-z}^2 + \frac{1-M}{2}\norm{x-z^+}^2.
\end{equation}
\end{itemize}
\end{proposition}
\begin{lemma}[Descent Lemma for smooth functions]
\label{prop:descent_lemma}
For $f$ proper differentiable and with a $L_f$-Lipschitz gradient, we have  $\forall x,y$
\begin{equation}
f(x) \leq f(y) + \langle \nabla f(y), x-y \rangle + \frac{L_f}{2}\norm{x-y}^2.
\end{equation}
\end{lemma}

\subsection{Proximal Gradient Descent  with a weakly convex function}
\label{sec:pgd}

We consider the following minimization problem for a smooth nonconvex function~$f$ and a weakly convex function $\phi$ that are both bounded from below:
\begin{equation}
    \label{eq:F}
    \min_x F(x):=\reglambda  f(x) + \phi(x).
\end{equation}

\noindent We now show under which conditions the classical Proximal Gradient Descent
\begin{equation}
    \label{eq:PGD}
    x_{k+1} \in \prox_{\tau \phi} \circ (\id - \tau\lambda \nabla f)(x_k)
\end{equation}
converges to a stationary point of~\eqref{eq:F}. 
We first show convergence of function
values, and then convergence of the iterates, if $F$ verifies the Kurdyka-Lojasiewicz (KL) property~\cite{attouch2013convergence}. Large classes of functions, in particular all the proper, closed, semi-algebraic functions~\cite{attouch2010proximal} satisfy this property, which is, in practice, the case of all the functions considered in this analysis. 

\begin{theorem}[Convergence of PGD algorithm~\eqref{eq:PGD}]
\label{prop:PGD}
Assume $f$ and $\phi$ proper lsc, bounded from below with $f$ differentiable with $L_f$-Lipschitz gradient, and $\phi$ $M$-weakly convex. Then for $\tau < 2/({\lambda L_f+M})$, the iterates~\eqref{eq:PGD} verify 
 \begin{itemize}
        \item[(i)] $(F(x_k))$ monotonically decreases and converges.
        \item[(ii)] $\norm{x_{k+1}-x_k}$ converges to $0$ at rate $\min_{k \leq K} \norm{x_{k+1}-x_k} = \mathcal{O}(1/\sqrt{K})$
        \item[(iii)] All cluster points of the sequence $x_k$ are stationary points of $F$.
        \item[(iv)] If the sequence $x_k$ is bounded and if $F$ verifies the KL property at the cluster points of $x_k$, then  $x_k$
        converges, with finite length, to a stationary point of $F$.
    \end{itemize}
\end{theorem}

The proof follows standard arguments of the convergence analysis of the PGD in the nonconvex setting~\cite{beck2009fast,attouch2013convergence,ochs2014ipiano}.
We only demonstrate here the first point of the theorem, the rest of the proof is detailed in Appendix~\ref{app:proof_PGD}.
\begin{proof}[i]
Relation~\eqref{eq:PGD} leads to
$\frac{x_{k}- x_{k+1}}{\tau} - \lambda\nabla f(x_k) \in \partial  \phi(x_{k+1})$,
by definition of the proximal operator. 
As $\phi$ is $M$-weakly convex,  Proposition~\ref{prop:weaklyconvex} (ii) leads to
\begin{equation}
\label{eq:after_weak_convexity}
\phi(x_{k}) \hspace{-2pt}\geq\hspace{-2pt} \phi(x_{k+1})\hspace{-2pt} +\hspace{-2pt} \frac{\norm{x_{k}-x_{k+1}}^2\hspace{-5pt}}{\tau} \hspace{-2pt}+ \hspace{-2pt}\lambda \langle  \nabla f(x_k), x_{k+1}\hspace{-2pt}-x_{k} \rangle - \frac{M}{2} \norm{ x_{k}-x_{k+1}}^2\hspace{-5pt}.
\end{equation}
The descent Lemma~\ref{prop:descent_lemma} gives for $f$:
\begin{equation}
f(x_{k+1}) \leq f(x_{k}) +  \langle  \nabla f(x_k), x_{k+1}-x_{k} \rangle + \frac{L_f}{2} \norm{ x_{k}-x_{k+1}}^2.
\end{equation}
Combining both inequalities, for $F_{\lambda,\sigma}=\lambda f+\phi_\sigma$, we obtain
\begin{equation}
\label{eq:decrease_F}
 F(x_{k}) \geq  F(x_{k+1})+ \left(\frac1\tau -\frac{M + \lambda L_f}{2} \right)  \norm{x_{k}-x_{k+1}}^2.
\end{equation}
Therefore, if $\tau < 2/({M+\lambda L_f})
$,  $(F(x_k))$ is monotically deacreasing. As $F$ is assumed lower-bounded, $(F(x_k))$ converges.\qed
\end{proof}

\subsection{Prox-PnP Proximal Gradient Descent (Prox-PnP-PGD)}\label{sec:pgd_prox}

Equipped with the convergence of PGD, we can now study the convergence of \emph{Prox-PnP-PGD}, the PnP-PGD algorithm with plugged Proximal Denoiser~\eqref{eq:proximal_denoiser}:
\begin{equation}
    \label{eq:PnP-PGDbis}
     x_{k+1} = D_\sigma(\id - \lambda f)(x_k) = \prox_{\phi_{\sigma}}(\id - \lambda f)(x_k).
\end{equation}
This algorithm targets stationary points of the functional $F_{\lambda,\sigma}$ defined as:
\begin{equation}
    \label{eq:F_tau}
    F_{\lambda,\sigma} := \reglambda  f + \phi_\sigma.
\end{equation}

The following result, obtained from Theorem~\ref{prop:PGD},  improves~\cite{hurault2022proximal} using the fact that the potential $\phi_\sigma$ is not any nonconvex function but a weakly convex one. 
\begin{corollary}[Convergence of Prox-PnP-PGD~\eqref{eq:PnP-PGDbis}]\label{thm:PnP-PGD} Let $g_\sigma : \mathbb{R}^n \to \mathbb{R} \cup \{+\infty\}$ of class $\mathcal{C}^2$, coercive, with $L_{g_\sigma}$-Lipschitz gradient, $L_{g_\sigma}<1$, and ${D_{\sigma} := \id - \nabla g_\sigma}$.
Let $\phi_{\sigma}$ be defined from $g_\sigma$ as in Proposition~\ref{prop:proxdenoiser}.   
    Let $f : \mathbb{R}^n \to \mathbb{R} \cup \{+\infty\}$  differentiable with $L_f$-Lipschitz gradient.
    Assume $f$ and $D_\sigma$ respectively KL and semi-algebraic, and $f$ and $g_\sigma$ bounded from below. 
    Then, for $\reglambda L_f < (L_{g_\sigma}+2)/(L_{g_\sigma}+1)$, the iterates $x_k$ given by the iterative scheme~\eqref{eq:PnP-PGDbis} verify the convergence properties (i)-(iv) of Theorem~\ref{prop:PGD} for $F=F_{\lambda,\sigma}$.
\end{corollary}

The proof of this result is given in Appendix~\ref{app:proof_PnP-PGD}. It is a direct application of Theorem~\ref{prop:PGD} using $\tau=1$ and the fact that $\phi_\sigma$ defined in Proposition~\ref{prop:proxdenoiser} is $M=L_{g_\sigma}/(L_{g_\sigma}+1)$-weakly convex. 

By exploiting the weak convexity of $\phi_\sigma$, the convergence condition $\lambda L_f<1$  of~\cite{hurault2022proximal} is here replaced by  $\reglambda L_f < \frac{L_{g_\sigma}+2}{L_{g_\sigma}+1}$. Even if the bound is improved, we are still limited to regularization parameters satisfying $\lambda L_f<2$. In the next section, we propose a modification of the PGD algorithm to relax this constraint.

\section{PnP Relaxed Proximal Gradient Descent (PnP-$\alpha$PGD)}\label{sec4}
In this section, we study the convergence of a relaxed PGD algorithm applied to problems~\eqref{eq:pb} involving a smooth  convex 
function $f$ and a weakly convex function~$\phi$. Our objective is to design a convergent algorithm in which the proximal operator is only computable for $\tau=1$ and the data-fidelity term constraint is less restrictive than the bound $\tau < 2/(M+\lambda L_f)$ of Theorem~\ref{prop:PGD}.
\subsection{$\alpha$PGD algorithm}
We present our main result which concerns, for weakly convex functions $\phi$, the convergence of the following $\alpha$-relaxed PGD algorithm, defined for  $0<\alpha<1$ as
\begin{subequations} \label{eq:PGD2}
\begin{align}[left = \empheqlbrace\,]
q_{k+1} &= (1-\alpha) y_k + \alpha x_k   \label{eq:PGD2_q} \\
x_{k+1} &= \prox_{\tau \phi}(x_k -\tau \lambda\nabla f(q_{k+1}))  \label{eq:PGD2_x}\\
y_{k+1} &= (1-\alpha) y_k + \alpha  x_{k+1}. \label{eq:PGD2_y}
\end{align}
\end{subequations}  
Algorithm~\eqref{eq:PGD2} with $\alpha=1$ exactly corresponds to the PGD algorithm~\eqref{eq:PGD}.
This scheme is reminiscent of~\cite{tseng2008accelerated} (taking $\alpha=\theta_k$ and $\tau=\frac{1}{\theta_k L_f}$ in Algorithm~1 of~\cite{tseng2008accelerated}), which generalizes Nesterov-like accelerated proximal gradient methods \cite{beck2009fast,nesterov2013gradient}. 
See also~\cite{dvurechensky18a} for a variant with line-search.
As shown in~\cite{lan2018optimal}, there is a strong connection between the proposed algorithm~\eqref{eq:PGD2} and the Primal-Dual algorithm~\cite{CP11} with Bregman proximal operator~\cite{chambolle2016ergodic}. 
In the \emph{convex} setting,  
one can show that ergodic convergence is obtained with $\tau \lambda L_f > 2$ and small values $\alpha$. Convergence of a similar algorithm is also shown in~\cite{mollenhoff2015primal} for a $M$-semi convex $\phi$ and a $c>M$-strongly convex $f$. However, $\phi_\sigma$ is here nonconvex while $f$ is only convex, so that a new convergence result needs to be derived. 

\begin{theorem}[Convergence of $\alpha$PGD~\eqref{eq:PGD2}]\label{thm:alphaPGD}Assume $f$ and $\phi$ proper lsc, bounded from below,  $f$ convex differentiable with $L_f$-Lipschitz gradient and $\phi$ $M$-weakly convex. 
Then\footnote{As shown in the proof, a better bound can be found, but with little numerical gain.} for $\alpha\hspace{-2pt} \in\hspace{-3pt} (0,\hspace{-2pt}1\hspace{-1pt})$ and
$ \tau \hspace{-2pt}< \hspace{-2pt}\min \hspace{-2pt}\left(\hspace{-2pt} \frac{1}{\alpha \lambda L_f}\hspace{-2pt},\hspace{-3pt} \frac{\alpha}{M}\hspace{-3pt}\right)\hspace{-1pt}$,\,
the updates~\eqref{eq:PGD2} verify
\begin{itemize}
    \item[(i)] $F(y_k) + \frac{\alpha}{2\tau}\left(1-\frac1\alpha\right)^2 \norm{y_k - y_{k-1}}^2$ monotonically decreases and converges.
    \item[(ii)] $\norm{y_{k+1}-y_k}$ converges to $0$ at rate $\min_{k \leq K} \norm{y_{k+1}-y_k} = \mathcal{O}(1/\sqrt{K})$
    \item[(iii)] All cluster points of the sequence $y_k$ are stationary points of $F$.
\end{itemize}
\end{theorem}
The proof, given in Appendix~\ref{app:proof_PGD2}, relies on Lemma~\ref{prop:descent_lemma} and Proposition~\ref{prop:weaklyconvex}. It also requires the convexity of $f$. 

With this theorem, $\alpha$PGD is shown to verify convergence of the iterates and of the norm of the residual to $0$. Note that we do not have here the analog of Theorem~\ref{thm:PnP-PGD}(iv) on the iterates' convergence using the KL hypothesis. Indeed, as we detail in Appendix~\ref{app:precision_KL}, the nonconvex convergence analysis with KL functions from~\cite{attouch2013convergence} or~\cite{ochs2014ipiano} do not extend to our case. 

When $\alpha=1$, Algorithms~\eqref{eq:PGD2} and~\eqref{eq:PGD} are equivalent, but we get a slightly worse bound in Theorem~\ref{thm:alphaPGD} than in Theorem~\ref{prop:PGD} ($ \tau < \min \left( \frac{1}{\lambda L_f}, \frac{1}{M}\right)\leq \frac2{\lambda L_f+M}$). Nevertheless, when used with $\alpha<1$, we next show that the relaxed algorithm is more relevant in the perspective of PnP with proximal denoiser.

\subsection{Prox-PnP-$\alpha$PGD algorithm}

We can now study the Prox-PnP-$\alpha$PGD algorithm obtained by taking the proximal denoiser~\eqref{eq:proximal_denoiser} in the $\alpha$PGD algorithm~\eqref{eq:PGD2} 
\begin{subequations} \label{eq:PnP-PGD2} 
\begin{align}[left = \empheqlbrace\,]
q_{k+1} &= (1-\alpha) y_k + \alpha x_k   \label{eq:PnP-PGD2_q} \\
x_{k+1} &= D_\sigma(x_k  - \lambda\nabla f(q_{k+1}))  \label{eq:PnP-PGD2_x}\\
y_{k+1} &= (1-\alpha) y_k + \alpha  x_{k+1}  \label{eq:PnP-PGD2_y}
\end{align}
\end{subequations} 
This scheme targets the minimization of the functional $F_{\lambda,\sigma}$ given in~\eqref{eq:F_tau}.

\begin{corollary}[Convergence of Prox-PnP-$\alpha$PGD~\eqref{eq:PnP-PGD2}]
    \label{thm:PnP-PGD2}Let $g_\sigma : \mathbb{R}^n \to \mathbb{R} \cup \{+\infty\}$ of class $\mathcal{C}^2$, coercive, with $L_g<1$-Lipschitz gradient and ${D_{\sigma} := \id - \nabla g_\sigma}$.
    Let $\phi_{\sigma}$ be the $M={L_{g_\sigma}}/{(L_{g_\sigma}+1)}$-weakly convex function defined from $g_\sigma$ as in Proposition~\ref{prop:proxdenoiser}.
    Let $f$ be proper, convex and differentiable with $L_f$-Lipschitz gradient.
    Assume $f$ and $g_\sigma$ bounded from below. 
    Then, if $\lambda L_fM  < 1$ and for any $\alpha \in [0,1]$
    \begin{equation}
    \label{eq:PnP-PGD2-conditions}\small 
    M<\alpha < 1/(\lambda L_f)   
    \end{equation}
    the iterates $x_k$ given by the iterative scheme~\eqref{eq:PnP-PGD2} verify the convergence properties (i)-(iii) of Theorem~\ref{thm:alphaPGD} for $F=F_{\lambda, \sigma}$ defined in~\eqref{eq:F_tau}.

\end{corollary}
This PnP corollary is obtained by taking $\tau=1$ in Theorem~\ref{thm:alphaPGD} and using the 
$M=(L_{g_\sigma})/(L_{g_\sigma}+1)$-weakly convex potential $\phi_\sigma$ defined in Proposition~\ref{prop:proxdenoiser}.

The existence of $\alpha\in[0,1]$ satisfying relation~\eqref{eq:PnP-PGD2-conditions} is ensured as soon as $\lambda L_f M < 1$. As a consequence, when $M$ gets small (\emph{i.e} $\phi_\sigma$ gets "more convex") $\lambda L_f$ can get arbitrarily large. This is a major advance compared to Prox-PnP-PGD that was limited (Corollary~\ref{thm:PnP-PGD}) to $\lambda L_f < 2$ even for convex $\phi$ ($M=0$).
To further exploit this property, we now consider the relaxed denoiser $D^\gamma_\sigma$~\eqref{eq:relaxed_proximal_denoiser} that is associated to a function $\phi_\sigma^\gamma$ with a tunable weak convexity constant $M^\gamma$.

\begin{corollary}[Convergence of Prox-PnP-$\alpha$PGD with relaxed denoiser]\label{thm:PnP-PGD3}
Let $F^\gamma_{\lambda,\sigma} := \reglambda  f + \phi^\gamma_\sigma$, with the $M^\gamma=\frac{\gamma L_g}{\gamma L_g+1}$-weakly convex potential $\phi^\gamma_\sigma$ introduced in~\eqref{eq:relaxed_proximal_denoiser_phi} and $L_g<1$. Then, for $M^\gamma <\alpha < 1/(\lambda L_f)$, the iterates $x_k$ given by the Prox-PnP-$\alpha$PGD~\eqref{eq:PnP-PGD2} with $\gamma$-relaxed denoiser $D^\gamma_\sigma$ defined in~\eqref{eq:relaxed_proximal_denoiser} verify the convergence properties (i)-(iii) of Theorem~\ref{thm:alphaPGD} for $F=F^\gamma_{\lambda,\sigma}$.
\end{corollary}
Therefore, using the $\gamma$-relaxed denoiser $D^\gamma_\sigma = \gamma D_\sigma + (1-\gamma)\id$, the overall convergence condition on $\lambda$ is now 
$
    \lambda < \frac{1}{L_f} \frac{\gamma L_g}{ \gamma L_g + 1}$.  

Provided $\gamma$ gets small, $\lambda$ can be arbitrarily large.  
Small $\gamma$ means small amount of regularization brought by denoising at each step of the PnP algorithm. Moreover, for small $\gamma$, the targeted regularization function $\phi_\sigma^\gamma$ gets close to a convex function and it has already been observed that deep convex regularization can be sub-optimal compared to more flexible nonconvex ones \cite{cohen2021has}. Depending on the inverse problem, and on the necessary amount of regularization,  the choice of the couple $(\gamma,\lambda)$ will be of paramount importance for efficient restoration.

\section{Experiments}
The efficiency of the proposed Prox-PnP-$\alpha$PGD algorithm~\eqref{eq:PnP-PGD2} is now demonstrated on deblurring and super-resolution.  
For both applications, we consider a degraded observation   $y = A x^* + \nu\in \mathbb{R}^m$ of a clean image $x^*\in\mathbb{R}^n$ that is estimated by solving problem~\eqref{eq:pb} with $f(x)=\frac{1}{2}\norm{Ax-y}^2$. Its gradient  $\nabla f=A^T(Ax-y)$ is thus Lipschitz with constant $L_f =\norm{A^T A}_S$.  We use for evaluation and comparison the 68 images from the CBSD68 dataset, center-cropped to $n=256\times256$ and Gaussian noise with $3$ noise levels $\nu \in\{0.01,0.03,0.05\}$.

For \emph{deblurring}, the degradation operator $A=H$ is a convolution performed with circular boundary conditions. As in~\cite{zhang2017learning,hurault2021gradient,pesquet2021learning,zhang2021plug}, we consider the 8  real-world camera shake kernels of~\cite{levin2009understanding}, the $9 \times9$ uniform kernel and the $25 \times 25$ Gaussian kernel with standard deviation $1.6$.

For single image \emph{super-resolution} (SR), the low-resolution image $y \in \mathbb{R}^m$ is obtained from the high-resolution one $x \in \mathbb{R}^n$ via $y = SHx + \nu$ where $H \in \mathbb{R}^{n \times n}$ is the convolution with anti-aliasing kernel.  The matrix $S$ is the standard $s$-fold downsampling matrix of size $m\times \dime$ and $\dime = s^2 \times m$.     As in~\cite{zhang2021plug}, we evaluate SR performance on 4 isotropic Gaussian blur kernels with  standard deviations $0.7$, $1.2$, $1.6$ and~$2.0$;  and consider downsampled images at scale $s=2$ and $s=3$.

The proximal denoiser $D_\sigma$ defined in Proposition~\ref{prop:proxdenoiser} is trained following~\cite{hurault2022proximal} with $L_g < 1$. For both Prox-PnP-PGD and Prox-PnP-$\alpha$PGD algorithm, we use the $\gamma$-relaxed version of the denoiser~\eqref{eq:relaxed_proximal_denoiser}. 
All the hypotheses on $f$ and $g_\sigma$ from Corollaries~\ref{thm:PnP-PGD} and~\ref{thm:PnP-PGD3} are thus verified and convergence of Prox-PnP-PGD and Prox-PnP-$\alpha$PGD are theoretically guaranteed provided the corresponding conditions on $\lambda$ are satisfied. 
Hyper-parameters $\gamma\in [0,1]$, $\lambda$ and $\sigma$ are optimized via grid-search. 
In practice, we found that the same choice of parameters $\gamma$ and $\sigma$ are optimal for both PGD and $\alpha$PGD, with values depending on the amount of noise $\nu$ in the input image. We thus choose $\lambda \in [0,\lambda_{lim}]$ where for Prox-PnP-PGD $\lambda_{lim}^{\text{PGD}} = \frac{1}{L_f}\frac{\gamma + 2}{\gamma + 1}$ and for Prox-PnP-$\alpha$PGD $\lambda_{lim}^{\alpha \text{PGD}} = \frac{1}{L_f}\frac{\gamma + 1}{\gamma} \geq \lambda_{lim}^{\text{PGD}} $. For both $\nu = 0.01$ and $\nu = 0.03$, $\lambda$ is set to its maximal allowed value $\lambda_{lim}$. Prox-PnP-$\alpha$PGD is  expected to outperform Prox-PnP-PGD at these noise levels.
Finally, for Prox-PnP-$\alpha$PGD, $\alpha$ is set to its maximum possible value $1 / (\lambda L_f)$. 

We numerically compare in Table~\ref{tab:whole_results} the presented methods Prox-PnP-PGD (that improves~\cite{hurault2022proximal}) and Prox-PnP-$\alpha$PGD against three state-of-the-art deep PnP approaches: IRCNN~\cite{zhang2017learning}, DPIR~\cite{zhang2021plug}, and GS-PnP~\cite{hurault2021gradient}. Among them, only GS-PnP has convergence guarantees. Both IRCNN and DPIR use PnP-HQS, the PnP version of the Half-Quadratic Splitting algorithm, with well-chosen varying stepsizes. GS-PnP uses the gradient-step denoiser~\eqref{eq:gs_denoiser} in PnP-HQS. 

\begin{figure*}[!ht] \centering
    \captionsetup[subfigure]{justification=centering}
    \begin{subfigure}[b]{.24\textwidth}
            \centering
            \begin{tikzpicture}[spy using outlines={rectangle,blue,magnification=5,size=1.2cm, connect spies}]
            \node {\includegraphics[height=2.5cm]{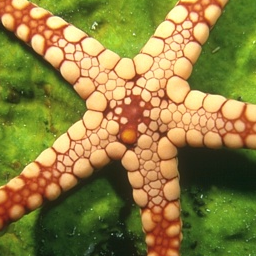}};
            \spy on (0.25,-0.4) in node [left] at  (1.25,.62);
                \node at (-0.85,-0.85) {\includegraphics[height=0.8cm]{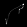}};
            \end{tikzpicture}\vspace*{-0.15cm}
            \caption{Clean \\ ~}
        \end{subfigure}
    \begin{subfigure}[b]{.24\textwidth}
            \centering
            \begin{tikzpicture}[spy using outlines={rectangle,blue,magnification=5,size=1.2cm, connect spies}]
            \node {\includegraphics[height=2.5cm]{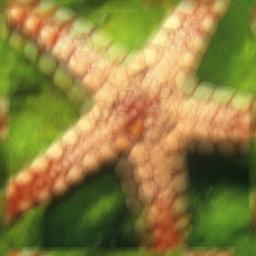}};
            \spy on (0.25,-0.4) in node [left] at  (1.25,.62);
            \end{tikzpicture}\vspace*{-0.15cm}
            \caption{Observed \\ ~}
        \end{subfigure}
    \begin{subfigure}[b]{.24\textwidth}
            \centering
            \begin{tikzpicture}[spy using outlines={rectangle,blue,magnification=5,size=1.2cm, connect spies}]
            \node {\includegraphics[width=2.5cm]{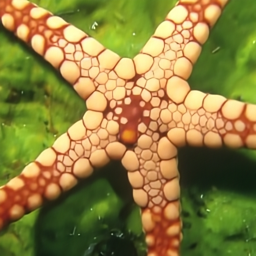}};
            \spy on (0.25,-0.4) in node [left] at (1.25,.62);
            \end{tikzpicture}\vspace*{-0.15cm}
            \caption{Prox-PnP-PGD ($33.32$dB) }
        \end{subfigure}
    \begin{subfigure}[b]{.24\textwidth}
            \centering
            \begin{tikzpicture}[spy using outlines={rectangle,blue,magnification=5,size=1.2cm, connect spies}]
            \node {\includegraphics[width=2.5cm]{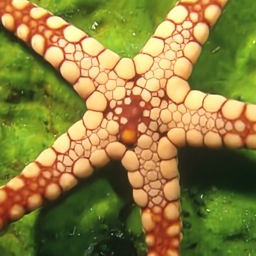}};
            \spy on (0.25,-0.4) in node [left] at (1.25,.62);
            \end{tikzpicture}\vspace*{-0.15cm}
            \caption{Prox-PnP-$\alpha$PGD ($33.62$dB)}
        \end{subfigure}
        \begin{subfigure}[b]{.26\textwidth}
            \centering
            \includegraphics[height=2.1cm]{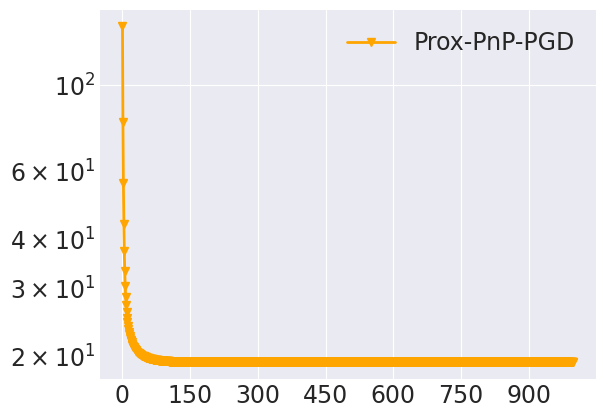}\vspace*{-0.15cm}
            \caption{$F_{\lambda,\sigma}(x_k)$ \\ Prox-PnP-PGD}
        \end{subfigure}
        \begin{subfigure}[b]{.26\textwidth}
            \centering
            \includegraphics[height=2.1cm]{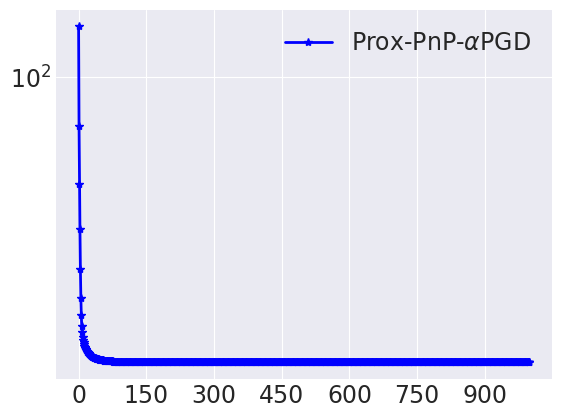}\vspace*{-0.15cm}
            \caption{$F_{\lambda,\sigma}(x_k)$ \\ Prox-PnP-$\alpha$PGD}
        \end{subfigure}
        \begin{subfigure}[b]{0.3\textwidth}
            \centering
            \includegraphics[width=3.5cm]{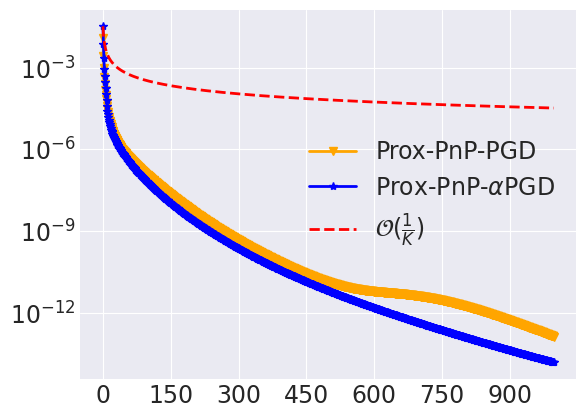}\vspace*{-0.15cm}
            \caption{\footnotesize $\norm{x_{i+1}-x_i}^2$}
        \end{subfigure}
    \caption{Deblurring of ``starfish" blurred with the shown kernel and noise $\nu=0.01$ .}
    \label{fig:deblurring}
    \end{figure*}

\begin{figure*}[!ht] \centering
    \captionsetup[subfigure]{justification=centering}
    \centering
    \begin{subfigure}[b]{.24\textwidth}
            \centering
            \begin{tikzpicture}[spy using outlines={rectangle,blue,magnification=5,size=1.5cm, connect spies}]
            \node {\includegraphics[width=2.5cm]{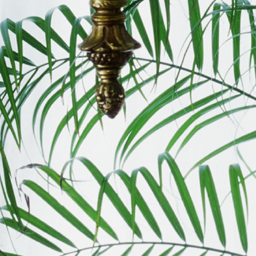}};
             \spy on (-0.4,0.4) in node [left] at (1.25,-0.55);
            \node at (-0.85,-0.85) {\includegraphics[width=0.8cm]{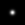}};
            \end{tikzpicture}\vspace*{-0.15cm}
            \caption{Clean\\ ~}
        \end{subfigure}
    \begin{subfigure}[b]{.24\textwidth}
            \centering
            \begin{tikzpicture}[spy using outlines={rectangle,blue,magnification=5,size=1.5cm, connect spies}]
            \node {\includegraphics[width=2.5cm]{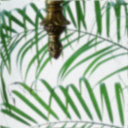}};
             \spy on (-0.4,0.4) in node [left] at (1.25,-0.55);
            \end{tikzpicture}\vspace*{-0.15cm}
            \caption{Observed \\ ~}
        \end{subfigure}
    \begin{subfigure}[b]{.24\textwidth}
            \centering
            \begin{tikzpicture}[spy using outlines={rectangle,blue,magnification=5,size=1.5cm, connect spies}]
            \node {\includegraphics[width=2.5cm]{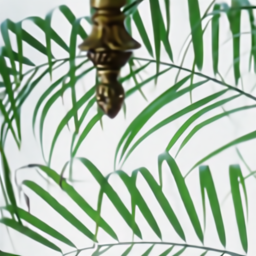}};
            \spy on (-0.4,0.4) in node [left] at (1.25,-0.55);
            \end{tikzpicture}\vspace*{-0.15cm}
            \caption{Prox-PnP-PGD ($28.19$dB)}
        \end{subfigure}
    \begin{subfigure}[b]{.24\textwidth}
            \centering
            \begin{tikzpicture}[spy using outlines={rectangle,blue,magnification=5,size=1.5cm, connect spies}]
            \node {\includegraphics[width=2.5cm]{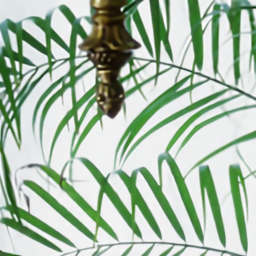}};
            \spy on (-0.4,0.4) in node [left] at (1.25,-0.55);
            \end{tikzpicture}\vspace*{-0.15cm}
            \caption{Prox-PnP-$\alpha$PGD ($28.59$dB)}
        \end{subfigure}
    \begin{subfigure}[b]{.26\textwidth}
            \centering
            \includegraphics[height=2.1cm]{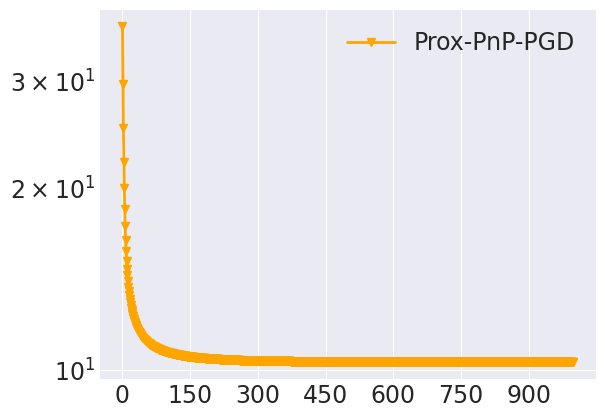}\vspace*{-0.15cm}
            \caption{$F_{\lambda,\sigma}(x_k)$ \\ Prox-PnP-PGD}
    \end{subfigure}
    \begin{subfigure}[b]{.26\textwidth}
            \centering
            \includegraphics[height=2.1cm]{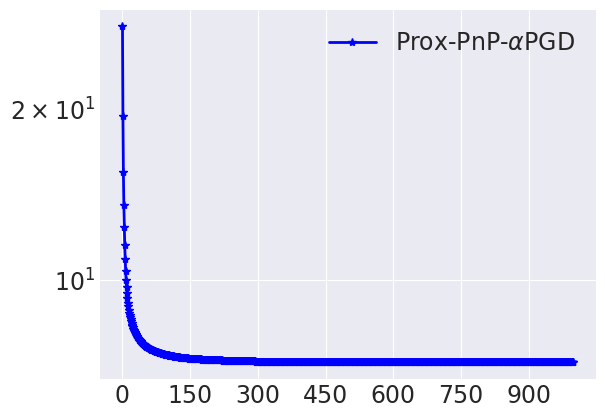}\vspace*{-0.15cm}
            \caption{$F_{\lambda,\sigma}(x_k)$ \\ Prox-PnP-$\alpha$PGD}
    \end{subfigure}
    \begin{subfigure}[b]{.3\textwidth}
            \centering
            \includegraphics[width=3.5cm]{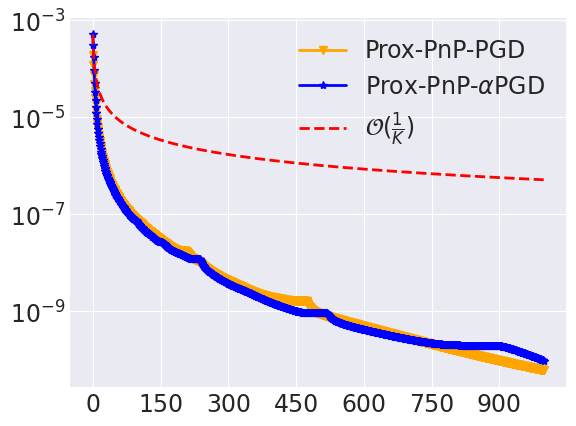}\vspace*{-0.15cm}
            \caption{$\norm{x_{i+1}-x_i}^2$}
    \end{subfigure}
    \caption{SR of "leaves" downscaled with the shown kernel, scale $2$ and $\nu=0.01$.}
    \label{fig:SR}
    \end{figure*}

As expected, by allowing larger values for $\lambda$, we observe that Prox-PnP-$\alpha$PGD  outperforms Prox-PnP-PGD in PSNR at low noise level $\nu \in \{0.01,0.03\}$. The performance gap is significant for deblurring and super-resolution with scale $2$ and $\nu = 0.01$, in which case only a low amount of regularization is necessary, that is to say a large $\lambda$ value. In these conditions, Prox-PnP-$\alpha$PGD almost closes the PSNR gap between Prox-PnP-PGD and the state-of-the-art PnP methods DPIR and GS-PnP that do not have any restriction on the choice of $\lambda$. We assume that the remaining difference of performance is due to the $\gamma$-relaxation of the denoising operation that affects the regularizer $\phi_\gamma$. 
We qualitatively verify in Figure~\ref{fig:deblurring} (deblurring) and Figure~\ref{fig:SR} (super-resolution), on "starfish" and "leaves" images, the performance gain of Prox-PnP-$\alpha$PGD over Prox-PnP-PGD. We also plot the evolution of $F_{\lambda,\sigma}(x_k)$ and $\norm{x_{k+1}-x_k}^2$ to empirically validate the convergence of both algorithms.

\begin{table}[h]
\centering\footnotesize\setlength\tabcolsep{3pt}
\begin{tabular}{c c c c  c c c c c c }
& \multicolumn{3}{c}{Deblurring}& \multicolumn{6}{c}{Super-resolution}\\
\cmidrule(lr){2-4}\cmidrule(lr){5-10}
& \multicolumn{3}{c}{}&\multicolumn{3}{c}{scale $s = 2$} & \multicolumn{3}{c}{scale $s = 3$} \\
\cmidrule(lr){5-7} \cmidrule(lr){8-10}%
    Noise level $\nu$ & 0.01 & 0.03 & 0.05 & 0.01 & 0.03 & 0.05 & 0.01 & 0.03 & 0.05 \\
    \midrule
    IRCNN~\cite{zhang2017learning}& $31.42$ & $28.01$ & $26.40$  & $26.97$ & $25.86$ & $25.45$ & $ 25.60$ & $ 24.72$ & $24.38$\\
    DPIR~\cite{zhang2021plug} & $\mathbf{31.93}$ & $\mathbf{28.30}$ & $\underline{26.82}$&$27.79$ & $26.58$ & $\underline{25.83}$ & $\mathbf{26.05}$ & $\underline{25.27}$ & $\underline{24.66}$ \\
   GS-PnP~\cite{hurault2021gradient} & $\underline{31.70}$ & $\underline{28.28}$ & $\mathbf{26.86}$  & $\underline{27.88}$ & $\mathbf{26.81}$ & $\mathbf{26.01}$ & $25.97$ & $\mathbf{25.35}$ & $\mathbf{24.74}$\\
   \midrule
   Prox-PnP-PGD & $30.91$ & $27.97$ & $26.66$ & $27.68$ & $26.57$ & $25.81$ & $25.94$ & $25.20$ & $24.62$ \\
   Prox-PnP-$\alpha$PGD & $31.55$ & $28.03$ & $26.66$& $\mathbf{27.92}$ & $\underline{26.61}$ & $25.80$ & $\underline{26.03}$ & $25.26$ & $24.61$ \\
\end{tabular}
\caption{PSNR (dB) results on CBSD68 for  deblurring (left) and super-resolution (right). PSNR are averaged over $10$ blur kernels for deblurring (left) and $4$ blur kernels along various scales $s$ for super-resolution (right).
}
\label{tab:whole_results}
\end{table}

\section{Conclusion}
In this paper, we propose a new convergent plug-and-play algorithm built from a relaxed version of the Proximal Gradient Descent (PGD) algorithm. When used with a proximal denoiser, while the original PnP-PGD imposes restrictive conditions on the parameters of the problem, the proposed algorithms converge, with minor conditions, towards stationary points of a weakly convex functional.
We illustrate numerically the convergence and the efficiency of the method. 
    
\section*{Acknowledgements}
    This work was funded by the French ministry of research through a CDSN grant of ENS Paris-Saclay. This study has also been carried out with financial support from the French Research Agency through the PostProdLEAP and Mistic projects (ANR-19-CE23-0027-01 and ANR-19-CE40-005).
    A.C. thanks Juan Pablo Contreras for many interesting discussions about nonlinear accelerated descent algorithms.

\bibliographystyle{splncs04}
\bibliography{mybibliography}

\newpage

\appendix
\setcounter{page}{1}

\section{Proof of Proposition~\ref{prop:proxdenoiser} (Characterization of the Proximal Denoiser)}
\label{app:proxdenoiser}
\begin{proof}
This result is an extension of the Proposition 3.1 from \cite{hurault2022proximal}. Under the same assumptions as ours, it is shown, using the characterization of the Proximity operator from \cite{gribonval2020characterization}, that for $\hat \phi_\sigma : \mathbb{R}^n \to \mathbb{R}$ defined as 
\begin{equation}\hspace*{-1cm}
    \hat \phi_\sigma(x):=\left\{\begin{array}{ll}   g_\sigma({D_\sigma}^{-1}(x)))-\frac{1}{2} \norm{{D_\sigma}^{-1}(x)-x}^2  \ \  \text{if }\ x \in \Im(D_\sigma),\\
     +\infty \ \ \text{otherwise} \end{array}\right.
\end{equation} 
we have $D_\sigma = \prox_{\hat \phi_\sigma}$ and  $\forall x \in \mathbb{R}^n$,  $\hat \phi_\sigma(x) \geq g_\sigma(x)$.

We would like to show that $\hat \phi_\sigma$ is semiconvex. 
    However, as $dom(\hat \phi_\sigma) = Im(D_\sigma)$ is non necessarily convex, we can not obtain this result. We thus propose another choice of function  $\phi_\sigma$ such that $D_\sigma = \prox_{\phi_\sigma}$.
    As we suppose that $\nabla g_\sigma  = \id - D_\sigma$ is $L$ Lipschitz, $D_\sigma$ is $L+1$ Lipschitz. 
    By~\cite[Proposition 2]{gribonval2020characterization},
    we have $\forall x \in \mathcal{X}$,
    \begin{equation}
    D_\sigma(x) \in \prox_{\phi_\sigma}(x)
    \end{equation}
    for some proper lsc fonction $\phi_\sigma : \mathcal{X} \to \mathbb{R}^n$ such that $ x \to \phi_\sigma(x) + \left( 1 - \frac{1}{L+1}\right)\frac{\norm{x}^2}{2}$ is convex, i.e. such that $\phi_\sigma$ is $1 - \frac{1}{L+1} = \frac{L}{L+1}$ weakly convex.
    For all $y \in \mathbb{R}^n$, the function $ x \to \phi_\sigma(x) + \frac{1}{2}\norm{x-y}^2$ is then strongly convex and the proximal operator is single valued. Then $\forall x \in \mathcal{X}$, 
    \begin{equation}
     D_\sigma(x) = \prox_{\phi_\sigma}(x)  = \prox_{\hat \phi_\sigma}(x)
     \end{equation}
     By~\cite[Corollary 5]{gribonval2020characterization}, for any $\mathcal{C} \subset \Im(D_\sigma)$ polygonally connected, there is $K \in \mathbb{R}$ such that 
     $\phi_\sigma = \hat \phi_\sigma + K $ on $\mathcal{C}$.
     As $\mathcal{X}$ is convex, it is connected and, as $D_{\sigma}$ is continuous, $\Im(D_\sigma)$ is connected and open, thus polygonally connected. Therefore, $\exists K \in \mathbb{R}$ such that $ \phi_\sigma = \hat \phi_\sigma + K$ on  $\Im(D_\sigma)$.
\end{proof}

\section{Proof of Proposition~\ref{prop:weaklyconvex} (Three points inequality for weakly convex functions)}\label{app:3points}

\begin{proof}
(i) and (ii) follow from the fact that $\phi + \frac{M}{2}\norm{x}^2$ is convex. We now prove (iii).
Optimality conditions of the proximal operator $z^+\in\prox_\phi(z)$ gives 
\begin{equation}
    z - z^+ \in \partial \phi(z^+).
\end{equation}
Hence, by (ii), we have $\forall x$,
\begin{equation}
\phi(x) \geq \phi(z^+) + \langle  z - z^+, x-z^+ \rangle - \frac{M}{2}\norm{x-z^+}^2,
\end{equation}
and therefore, 
\begin{equation*}
\begin{split}
     \phi(x) + \frac{1}{2}\norm{x-z}^2
    & \geq \phi(z^+) + \frac{1}{2}\norm{x-z}^2 + \langle z - z^+, x-z^+ \rangle - \frac{M}{2}\norm{x-z^+}^2 \\
    &= \phi(z^+) + \frac{1}{2}\norm{x-z^+}^2 + \frac{1}{2}\norm{z-z^+}^2 - \frac{M}{2}\norm{x-z^+}^2 \\
    &= \phi(z^+)  + \frac{1}{2}\norm{z-z^+}^2 + \frac{1-M}{2}\norm{x-z^+}^2.
\end{split}
\end{equation*}\qed
\end{proof}

\section{Proof of Theorem~\ref{prop:PGD} (PGD for nonconvex and weakly convex optimization)}
\label{app:proof_PGD}
We here demonstrate the last three points of Theorem~\ref{prop:PGD}. \begin{proof}
\begin{itemize}
    \item[(i)] We have demonstrated Section~\ref{sec:pgd} that $(F(x_k))$ is monotically deacreasing and converges. We call $F^*$ its limit.
        \item[(ii)]
            Summing~\eqref{eq:decrease_F} over $k=0,1,...,m$ gives
            \begin{equation}
                \begin{split}
                    \sum_{k=0}^m \norm{x_{k+1}-x_k}^2 &\leq \frac{1}{\frac1\tau -\frac{M + L_f}{2} }\left(F(x_0) -F(x_{m+1})\right) \\
                    &\leq \frac{1}{\frac1\tau -\frac{M + L_f}{2} } \left(F(x_0)-F^*\right) .
                \end{split}
            \end{equation}
            Therefore, $\lim_{k\to\infty} \norm{x_{k+1}-x_k} = 0$ with the convergence rate 
            \begin{equation}
            \gamma_k = \min_{0 \leq i \leq k}\norm{x_{i+1}-x_i}^2 \leq \frac{1}{k}  \frac{F(x_0)-F^*}{\frac1\tau -\frac{M + L_f}{2} }
            \end{equation}
        
        \item[(iii)]
        
        Suppose that a subsequence $(x_{k_i})_i$ is converging towards $y$. Let's show that $x$ is a critical point of $F$.
        We had
        \begin{equation}
                \frac{x_{k+1}-x_k}{\tau} -\lambda\nabla f(x_{k+1}) \in \partial \phi(x_{k+1}).
        \end{equation}
         From the continuity of $\nabla f$, we have $ \nabla f (x_{k_i}) \to \nabla f (y)$. As $\norm{x_{k+1}-x_k}\to 0$, we get
        \begin{equation}
             \frac{x_{k_i}-x_{k_i-1}}{\tau} -\lambda \nabla f(x_{k_i}) \to -\lambda \nabla f (y).
        \end{equation}
        We recall that the subdifferential of a proper, nonconvex function $\phi$ is defined as the limiting subdifferential
        \begin{equation}
\begin{split}
\partial \phi(x):=\biggl\{ \biggr.&v \in \mathbb{R}^n, \exists x_k, \phi(x_k) \rightarrow \phi(x), v_k \rightarrow v,\\&\biggl.  \varliminf_{z \to x_k} \frac{\phi(z)-\phi(x_k) - \langle v_k,z-x_k \rangle }{\norm{z-x_k}} \geq 0 \ \forall k \biggr\}.
\end{split}
        \end{equation}
        which verifies
        \begin{equation}
            \label{eq:subdiff_prop}
            \left\{ v \in \mathbb{R}^n, \exists x_k, \phi(x_k) \rightarrow \phi(x), v_k \rightarrow v, v_k \in \partial \phi(x_k) \right\} \subseteq \partial \phi(x).
        \end{equation}
        Therefore, if we can show that $\phi(x_{k_i}) \rightarrow \phi(y)$, we get $ -\lambda \nabla f (y) \in \partial \phi(y)$
        i.e. $y$ is a critical point of $F$.
        Using the fact that $\phi$ is lsc, 
        \begin{equation}
            \liminf_{i \to \infty} \phi(x_{k_i})  \geq \phi(y).
        \end{equation}
        On the other hand, by weak convexity of $\phi$, Proposition~\ref{prop:weaklyconvex} (ii) gives, for 
        ${z_{k_i} =  \frac{x_{k_i-1}- x_{k_i}}{\tau} - \lambda\nabla f(x_{k_i}) \in \partial \phi(x_{k_i})}$, 
        \begin{equation} \
        \begin{split}
        & \phi(x_{k_i}) \leq \phi(y) - \langle z_{k_i}, x_{k_i} - y \rangle + \frac{M}{2}\norm{x_{k_i} - y }^2 \\
        &\leq \phi(y) + \norm{z_{k_i}} \norm{ x_{k_i} - y} + \frac{M}{2}\norm{x_{k_i} - y }^2 \\ 
        &\leq \phi(y) + \left( \frac{\norm{x_{k_i-1}- x_{k_i}}}{\tau} + \lambda\norm{\nabla f(x_{k_i}) }\hspace{-1pt} +\hspace{-1pt}  \frac{M}{2} \norm{ x_{k_i} - y} \right)\hspace{-1pt} \norm{ x_{k_i} - y}.
        \end{split}
        \end{equation}
Therefore,
        \begin{equation}
        \limsup_{i \to \infty} \phi(x_{k_i}) \leq \phi(y),
        \end{equation}
        and 
        \begin{equation}
        \lim_{i \to \infty} \phi(x_{k_i}) = \phi(y).
        \end{equation}
        \item[(iv)] 
        We wish to apply Theorem 2.9 from~\cite{attouch2013convergence}. We need to satisfy H1, H2, and~H3 specified in~\cite[Section 2.3]{attouch2013convergence}. Condition H1 corresponds to the sufficient decrease condition shown in (i). For condition H3, we use that, as $(x_k)$ is bounded, there exists a subsequence $(x_{k_i})$ converging towards $y$.
        Then $F(x_{k_i}) \to F(y)$ has been shown in (iii). Finally, for condition H2, from~\eqref{eq:after_weak_convexity}, we had that 
        \begin{equation}
        z_{k+1} +\lambda \nabla f(x_{k+1}) = \frac{x_{k+1} - x_k}{\tau}
        \end{equation}
        where $z_{k+1} \in \partial \phi(x_{k+1})$. Therefore, 
        \begin{equation}
        \norm{z_{k+1} + \lambda\nabla f(x_{k+1})} = \frac{\norm{x_{k+1} - x_k}}{\tau}
        \end{equation} which gives condition H2. \hfill\qed
        \end{itemize}
        \end{proof}

\section{Proof of Corollary~\ref{thm:PnP-PGD} (Prox-PnP-PGD convergence)}\label{app:proof_PnP-PGD}

\begin{proof}
The convergence results is obtained applying Theorem~\ref{prop:PGD} for $\phi = \phi_\sigma$ defined in Proposition~\ref{prop:proxdenoiser} and $\tau = 1$. 
The condition $\tau \lambda L_f + M < 2$ from Theorem~\ref{prop:PGD}  becomes, with $\tau=1$ and the fact that $\phi_\sigma$ is $M=L_{g_\sigma}/(L_{g_\sigma}+1)$-weakly convex
\begin{equation}
    \lambda L_f < 2 - M = \frac{L_{g_\sigma}+2}{L_{g_\sigma}+1}
\end{equation}
Point (iv) requires $F_{\lambda,\sigma}$ to verify the Kurdyka-Lojasiewicz (KL) property at the cluster points of $x_k$, and the iterates $x_k$ to be bounded. We now show these two points. 
\begin{itemize}
    \item As $f$ is supposed KL, we only need to show that $\phi_\sigma$ is KL on the open $Im(D_\sigma)$. Indeed, cluster points of $x_k$ are fixed points of the PGD operator $D_\sigma \circ (\id - \tau \nabla f)$ and thus belong to $Im(D_\sigma)$. We supposed $D_\sigma$ to be a semi-algebraic mapping. As mentioned in \cite{attouch2013convergence}, by the Tarski–Seidenberg theorem, the composition and inverse of semi-algebraic mappings are semi-algebraic mappings.  Moreover, semi-algebraic scalar functions are KL. Therefore, by the definition of $\hat \phi_\sigma$ equation \eqref{eq:phi}, we get that $\phi_\sigma$ is semi-algebraic and thus KL on $Im(D_\sigma)$.
    
    Let us precise here that, in practice, $D_\sigma$ will be defined, as in \cite{hurault2022proximal}, as the gradient of a neural network. It is easy to check, using the chain rule and the fact that semi-algebraicity is stable by product, sum, inverse and composition that $D_\sigma$ is a then a semi-algebraic mapping. 
\item Using Proposition~\ref{prop:proxdenoiser} with the notations of the same Proposition, we have that $\forall x \in \mathbb{R}^n$, $g_\sigma(x) \leq \hat \phi_\sigma(x)$ Thus, the coercivity of $g_\sigma$ implies the coercivity of $\hat \phi_\sigma$. Moreover $\forall x \in Im(D_\sigma)$,
$\phi_\sigma = \hat \phi_\sigma + K$. 
Therefore, as $x_k \in Im(D_\sigma)$, we have along the sequence that 
$F_{\lambda,\sigma} = \lambda f(x_k) + \phi_\sigma(x_k) = \lambda f(x_k) + \hat \phi_\sigma + K$. Therefore, by coercivity of $\hat \phi_\sigma$ (and lower boundedness of $f$) and the fact that $F_{\lambda,\sigma}(x_k)$ monotonically decreases, the sequence $x_k$  remains necessarily bounded.

\end{itemize}

\end{proof}
\section{Proof of Theorem~\ref{thm:PnP-PGD2} ($\alpha$PGD for convex and weakly convex optimization)}
\label{app:proof_PGD2}

\begin{proof}
(i) and (ii) : 
We can write~\eqref{eq:PGD2_x} as 
\begin{equation}
\begin{split}
     x_{k+1} &\in \argmin_{y \in \mathbb{R}^n} \phi(y) +\lambda \langle \nabla f(q_{k+1}),y-x_k \rangle + \frac{1}{2\tau} \norm{y-x_k}^2 \\
     &\in \argmin_{y \in \mathbb{R}^n} \phi(y) + \lambda f(q_{k+1}) + \lambda\langle \nabla f(q_{k+1}),y-q_{k+1} \rangle + \frac{1}{2\tau} \norm{y-x_k}^2 \\
     &\in \argmin_{y \in \mathbb{R}^n} \Phi(y) +\frac{1}{2\tau} \norm{y-x_k}^2
\end{split}
\end{equation}
with $\Phi(y) := \phi(y) + \lambda f(q_{k+1}) +\lambda \langle \nabla f(q_{k+1}),y-q_{k+1}\rangle$.  
As $\phi$ is $M$-weakly convex, so does $\Phi$.
The three-points inequality of Proposition~\ref{prop:weaklyconvex} (iii) applied to $\Phi$ thus gives  $\forall y \in\mathbb{R}^n$,
\begin{equation}
\Phi(y) + \frac{1}{2\tau} \norm{y-x_k}^2 \geq \Phi(x_{k+1}) + \frac{1}{2\tau}\norm{ x_{k+1}-x_k}^2 + \left(\frac{1}{2\tau} - \frac{M}{2} \right) \norm{x_{k+1}-y}^2
\end{equation}
that is to say,
\begin{equation}
\begin{split}
\label{eq:after_3point}
 & \phi(y) + \lambda f(q_{k+1}) + \lambda \langle \nabla f(q_{k+1}),y-q_{k+1} \rangle
  +\frac{1}{2\tau} \norm{y-x_k}^2 \geq \\
 &\phi(x_{k+1}) + \lambda f(q_{k+1}) + \lambda \langle \nabla f(q_{k+1}),x_{k+1}-q_{k+1} \rangle \\
 &+ \frac{1}{2\tau}\norm{x_{k+1}-x_k}^2  + \left(\frac{1}{2\tau} - \frac{M}{2} \right) \norm{x_{k+1}-y}^2 
 \end{split}
\end{equation}
Using relation~\eqref{eq:PGD2_y}, and the descent Lemma~\ref{prop:descent_lemma} as well as the convexity on $f$,
\begin{equation}\label{relation2}
    \begin{split}
        & f(q_{k+1}) + \langle \nabla f(q_{k+1}),x_{k+1}-q_{k+1} \rangle \\
        =& f(q_{k+1}) +  \left\langle \nabla f(q_{k+1}), \frac{1}{\alpha} y_{k+1} + \left(1-\frac{1}{\alpha}\right) y_{k} - q_{k+1} \right\rangle \\
        =&\frac{1}{\alpha} \Big( f(q_{k+1}) + \langle \nabla f(q_{k+1}),y_{k+1} - q_{k+1} \rangle \Big)\\  &+
        \left(1-\frac{1}{\alpha}\right)\Big( f(q_{k+1}) + \langle \nabla f(q_{k+1}),y_{k}- q_{k+1} \rangle \Big) \\
        \geq &\frac{1}{\alpha} \left(  f(y_{k+1}) - \frac{L_f}{2} \norm{y_{k+1}-q_{k+1}}^2 \right)\\& +
        \left(1-\frac{1}{\alpha}\right) \Big( f(q_{k+1}) + \langle \nabla f(q_{k+1}),y_{k}- q_{k+1} \rangle \Big) \\
        \geq &\frac{1}{\alpha} \left(  f(y_{k+1}) - \frac{L_f}{2}\norm{y_{k+1}-q_{k+1}}^2 \right) +        \left(1-\frac{1}{\alpha}\right)f(y_k). 
\end{split}
\end{equation}
Since $y_{k+1}-q_{k+1} = \alpha(x_{k+1}-x_{k})$ (from relations~\eqref{eq:PGD2_q} and~\eqref{eq:PGD2_y}), by combining relations~\eqref{eq:after_3point} and~\eqref{relation2}, we now have for all $y \in\mathbb{R}^n$,
\begin{equation}
    \begin{split}
     & \phi(y) +  \lambda\left(\frac{1 }{\alpha}-1\right) f(y_k) + \frac{1}{2\tau} \norm{y-x_k}^2 + \lambda f(q_{k+1}) + \lambda \langle \nabla f(q_{k+1}),y-q_{k+1} \rangle \geq \\
     &\phi(x_{k+1})\hspace{-1pt} + \hspace{-1pt} \frac{\lambda}{\alpha} f(y_{k+1})\hspace{-1pt} +\hspace{-2pt} \left(\frac{1}{2\tau} \hspace{-1pt}-\hspace{-1pt} \frac{\alpha \lambda L_f}{2}\right) \norm{x_{k+1}-x_k}^2 \hspace{-2pt} +\hspace{-1pt}\left(\hspace{-1pt}\frac{1}{2\tau}\hspace{-1pt}- \hspace{-1pt}\frac{M}{ 2 }\right) \hspace{-2pt}\norm{x_{k+1}-y}^2.
     \end{split}
    \end{equation}
Using again the convexity of $f$ we get for all $y \in\mathbb{R}^n$,
\begin{equation}\label{eq:relation1}
    \begin{split}
     & \phi(y) +  \lambda f(y)  + \lambda \left(\frac{1}{\alpha}-1\right) f(y_k) + \frac{1}{2\tau} \norm{y-x_k}^2 \geq 
      \\
     &\phi(x_{k+1})\hspace{-1pt} + \hspace{-2pt} \frac{\lambda }{\alpha} f(y_{k+1}) + \left(\frac{1}{2\tau} - \frac{\alpha \lambda L_f}{2}\hspace{-1pt}\right) \norm{x_{k+1}-x_k}^2  \hspace{-2pt}+\hspace{-1pt} \left(\frac{1}{2\tau}- \frac{M}{ 2 }\right) \norm{x_{k+1}-y}^2.
     \end{split}
    \end{equation}
Now, the weak convexity of $\phi$ with relation~\eqref{eq:PGD2_y} gives 
\begin{equation}
\label{eq:phiconvex}
\phi(x_{k+1}) \geq \frac{1}{\alpha} \phi(y_{k+1}) + \left(1-\frac{1}{\alpha}\right) \phi(y_k) - \frac{M}{2}(1-\alpha) \norm{y_k - x_{k+1}}^2.
\end{equation}
Combining~\eqref{eq:relation1} and~\eqref{eq:phiconvex},  and using $F = \lambda f + \phi$ leads to
\begin{equation}
\begin{split}
       \label{eq:descent_semiconvex}
    \forall y \in\mathbb{R}^n 
    &\left(\frac{1}{\alpha}-1\right) (F(y_k) - F(y)) + \frac{1}{2\tau} \norm{y-x_k}^2 \geq \\
     &\frac{1}{\alpha} (F(y_{k+1}) - F(y)) + \left(\frac{1}{2\tau} - \frac{\alpha \lambda L_f}{2}\right) \norm{x_{k+1}-x_k}^2  \\ &+ \left(\frac{1}{2\tau}- \frac{M}{ 2 }\right) \norm{x_{k+1}-y}^2 
     - \frac{M}{2}(1-\alpha) \norm{y_k - x_{k+1}}^2.
\end{split}
\end{equation}
For $y=y_k$, we get 
\begin{equation}
\begin{split}
   \frac{1}{\alpha} (F(y_{k}) - F(y_{k+1})) &\geq 
   - \frac{1}{2\tau} \norm{y_k-x_k}^2 
     + \left(\frac{1}{2\tau} - \frac{\alpha \lambda L_f}{2}\right) \norm{x_{k+1}-x_k}^2  \\ &+ \left(\frac{1}{2\tau}- \frac{M(2-\alpha)}{ 2 }\right) \norm{x_{k+1}-y_k}^2.
\end{split}
\end{equation}
For constant $\alpha \in (0,1)$, using that 
\begin{align}
    & y_k - x_{k+1} = \frac{1}{\alpha}(y_k - y_{k+1}) \\
    & y_{k} - x_{k} = \left(1-\frac{1}{\alpha}\right)(y_k - y_{k-1})
\end{align}
we get 
\begin{equation}
\label{eq:before_cases}
  \begin{split}
   F(y_{k}) - F(y_{k+1}) \geq& - \frac{\alpha}{2\tau}\left(1-\frac1\alpha\right)^2 \norm{y_k-y_{k-1}}^2 \\ & 
     + \alpha \left(\frac{1}{2\tau} - \frac{\alpha \lambda L_f}{2}\right) \norm{x_{k+1}-x_k}^2   \\ &+ \frac1\alpha \left(\frac{1}{2\tau}- \frac{M(2-\alpha)}{ 2 }\right) \norm{y_{k+1}-y_k}^2.
     \end{split}
\end{equation}
With the assumption $\tau \alpha \lambda L_f < 1$, the second term of the right-hand side is non-negative and therefore,
\begin{equation} \label{eq:descent_aPGD}
  \begin{split}
  F(y_{k}) - F(y_{k+1}) &\geq - \frac{\alpha}{2\tau}\left(1-\frac1\alpha\right)^2 \norm{y_k-y_{k-1}}^2 \\ &+ \frac1\alpha \left(\frac{1}{2\tau}- \frac{M(2-\alpha)}{ 2 } \right) \norm{y_{k+1}-y_k}^2. \\
     &= -\delta \norm{y_k - y_{k-1}}^2 + \delta \norm{y_{k+1} - y_{k}}^2 \\ &+ (\gamma-\delta) \norm{y_{k+1} - y_{k}}^2
    \end{split}
    \end{equation}
with 
\begin{align}
    \delta &= \frac{\alpha}{2\tau}\left(1-\frac1\alpha\right)^2 \\
    \gamma &=  \frac1\alpha \left(\frac{1}{2\tau}- \frac{M(2-\alpha)}{ 2 }\right)
\end{align}

We now make use of the following lemma. 
\begin{lemma}[\cite{bauschke2011convex}]
\label{lem:an_bn}
Let $(a_n)_{n \in \mathbb{N}}$ and $(b_n)_{n \in \mathbb{N}}$ be two real sequences such that $b_n \geq 0$ $\forall n \in \mathbb{N}$, $(a_n)$ is bounded from below and $a_{n+1}+b_n \leq a_n$ $\forall n \in \mathbb{N}$. Then $(a_n)_{n \in \mathbb{N}}$ is a monotonically non-increasing and convergent sequence and $\sum_{n \in \mathbb{N}} b_n < +\infty$
\end{lemma}
To apply Lemma~\ref{lem:an_bn} we look for a stepsize satisfying 
\begin{equation}
    \gamma - \delta > 0 \quad \text{i.e.} \quad \tau < \frac{\alpha}{M}.
\end{equation}
Therefore, hypothesis $\tau < \min \left(\frac{1}{\alpha \lambda L_f},\frac{\alpha}{M}\right)$ gives that $(F(y_k) + \delta \norm{y_k - y_{k-1}}^2)$  is a non-increasing and convergent sequence and that $\sum_k \norm{y_k - y_{k+1}}^2 < + \infty$.

Note that a slightly more precise bound can be found keeping the second term in~\eqref{eq:before_cases}. For sake of completeness, we develop it here. Keeping the assumption $\tau \alpha \lambda L_f < 1$ in~\eqref{eq:before_cases}. We can use that
 \begin{equation}
 x_{k+1}-x_{k} = \frac{1}{\alpha}(y_{k+1}-y_k) + \left(1-\frac{1}{\alpha}\right)(y_{k}-y_{k-1}).
\end{equation}
Then, by convexity of the squared $\ell_2$ norm, for $0 < \alpha < 1$, we have 
\begin{equation}
\norm{y_{k+1}-y_k}^2 \leq \alpha \norm{x_{k+1}-x_{k}}^2 + (1-\alpha) \norm{y_{k}-y_{k-1}}^2
\end{equation}
and  
\begin{equation}\label{tmp}
\norm{x_{k+1}-x_{k}}^2 \geq \frac{1}{\alpha}\norm{y_{k+1}-y_k}^2  +  \left(1-\frac{1}{\alpha}\right) \norm{y_{k}-y_{k-1}}^2.
\end{equation}
which gives finally
\begin{equation}
\begin{split}
  & F(y_{k}) - F(y_{k+1}) \geq \\ &\left(\alpha\left(1-\frac1\alpha\right) \left(\frac{1}{2\tau} - \frac{\alpha L_f}{2}\right)  - \frac{\alpha}{2\tau}\left(1-\frac1\alpha\right)^2 \right) \norm{y_k-y_{k-1}}^2 \\
      &+ \left( \frac{1}{2\tau} - \frac{\alpha \lambda L_f}{2} + \frac1\alpha (\frac{1}{2\tau}- \frac{M(2-\alpha)}{ 2 }) \right)
      \norm{y_{k+1}-y_k}^2. \\
    &=  - \frac{1-\alpha}{2\alpha\tau} \left( 1 - \alpha^2 \tau  \lambda L_f \right) \norm{y_k-y_{k-1}}^2 \\
    &+ \frac{1}{2\alpha\tau} \left( 1 + \alpha - \alpha^2\tau \lambda L_f - \tau M(2-\alpha) \right)
      \norm{y_{k+1}-y_k}^2. \\
    &= -\delta \norm{y_k - y_{k-1}}^2 + \delta \norm{y_{k+1} - y_{k}}^2 + (\gamma-\delta) \norm{y_{k+1} - y_{k}}^2
    \end{split}
    \end{equation}
with 
\begin{align}
    \delta &= \frac{1-\alpha}{2\alpha\tau} \left( 1 - \alpha^2 \tau \lambda L_f \right) \\
    \gamma &=  \frac{1}{2\alpha\tau} \left( 1 - \alpha^2\tau \lambda L_f + \alpha - \tau M(2-\alpha) \right)
\end{align}
The condition on the stepsize becomes
\begin{equation}
\begin{split}
    & \gamma - \delta > 0 \\
    &\Leftrightarrow \tau < \frac{2\alpha}{\alpha^3 L_f + (2-\alpha)M}
\end{split}
\end{equation}
And the overall condition is 
\begin{equation}
     \tau < \min \left( \frac{1}{\alpha L_f}, \frac{2\alpha}{\alpha^3 L_f + (2-\alpha)M}\right)
\end{equation}

(iii) The  proof of this result is an extension of the proof proposed in Appendix~\ref{app:proof_PGD} in the context of the classical PGD. 
Suppose that a subsequence $(y_{k_i})$ is converging towards $y$. Let's show that $y$ is a critical point of $F$.
From~\eqref{eq:PGD2_x}, we have 
\begin{equation}
        \frac{x_{k+1}-x_k}{\tau} -\lambda \nabla f(q_{k+1}) \in \partial \phi (x_{k+1}).
\end{equation}
First we show that $x_{k_i+1}-x_{k_i} \to 0$. We have $\forall k > 1 $,
\begin{equation}
\begin{split}
\norm{x_{k+1}-x_k} &= \norm{ \frac{1}{\alpha} y_{k+1} + (1-\frac{1}{\alpha})y_{k} - \frac{1}{\alpha}y_{k} - (1-\frac{1}{\alpha})y_{k-1} } \\
&\leq  \frac{1}{\alpha}\norm{y_{k+1}-y_k} +  (\frac{1}{\alpha}-1)\norm{y_{k}-y_{k-1}} \\
&\to 0.
\end{split}
\end{equation}
From $\eqref{eq:PGD2_q}$, we also get $\norm{q_{k+1}-q_k} \to 0$. Now, let's show that $x_{k_i} \to y$ and $q_{k_i} \to y$. First using $\eqref{eq:PGD2_y}$, we have
\begin{equation}
\begin{split}
\norm{x_{k_i}-y} &\leq \norm{x_{k_i+1}-y}  + \norm{x_{k_i+1}-x_{k_i}} \\
&\leq \frac{1}{\alpha}\norm{y_{k_i+1}-y} + (\frac{1}{\alpha}-1)\norm{y_{k_i}-y}+ \norm{x_{k_i+1}-x_{k_i}} \\
&\to 0.
\end{split}
\end{equation}
Second, from $\eqref{eq:PGD2_q}$, we get in the same way $q_{k_i} \to y$. From the continuity of $\nabla f$, we get $ \nabla f (q_{k_i}) \to \nabla f (y)$ and therefore 
\begin{equation}
     \frac{x_{k_i}-x_{k_i-1}}{\tau} - \lambda \nabla f(q_{k_i}) \to - \lambda \nabla f (y).
\end{equation}
As explained in the proof Appendix~\ref{app:proof_PGD} (iii), 
if we can also show that $\phi(x_{k_i}) \rightarrow \phi(y)$, we get from the subdifferential characterization~\eqref{eq:subdiff_prop} that $ - \lambda \nabla f (y) \in \partial \phi(y)$ i.e. $y$ is a critical point of $F$.

Using the fact that $\phi$ is lsc and $x_{k_i} \to y$. 
\begin{equation}
    \liminf_{i \to \infty} \phi(x_{k_i})  \geq \phi(y).
\end{equation}
On the other hand, with Equation \eqref{eq:relation1} for $k+1 = k_i$, taking $i \to + \infty$, $\norm{y-x_{k_i+1}} \to 0$,  $\norm{y-x_{k_i}} \to 0$, $f(y_{k_i}) \to f(y)$,  $f(y_{k_i+1}) \to f(y)$ and we get 
\begin{equation}
\limsup_{i \to \infty} \phi(x_{k_i}) \leq \phi(y),
\end{equation}
and therefore
\begin{equation}
\lim_{i \to \infty} \phi(x_{k_i}) = \phi(y).
\end{equation}

\subsubsection{On the convergence of the iterates with the KL hypothesis}
\label{app:precision_KL}

In order to prove a result similar to Theorem~\ref{thm:PnP-PGD} (iv) on the convergence of the iterates with the KL hypothesis, we can not directly apply Theorem 2.9 from~\cite{attouch2013convergence} on $F$ as the objective function $F(x_k)$ by itself does not decrease along the sequence but $F(x_k) + \delta \norm{x_{k+1}-x_k}^2$ does (where $\delta = \frac{\alpha}{2\tau}\left(1-\frac1\alpha\right)^2$). 

Our situation is more similar to the variant of this result presented in~\cite[Theorem 3.7]{ochs2014ipiano}. 
Indeed, denoting $\mathcal{F} : \mathbb{R}^n \times \mathbb{R}^n \to  \mathbb{R}$ defined as $\mathcal{F}(x,y) = F(x) + \delta \norm{x-y}^2$ and considering $\forall k \geq 1$, the sequence $z_k = (y_k,y_{k-1})$ with $y_k$ following our algorithm, we can easily show that $z_k$ verifies the conditions H1 and H3 specified in~\cite[Section 3.2]{ochs2014ipiano}. However, condition H2 does not extend to our algorithm. We plan as future work to derive a new version of \cite[Section 3.2]{ochs2014ipiano} that fits to our case of interest.

\end{proof}

\end{document}